%% file: main.tex
\documentclass{article}
\usepackage{spconf,amsmath,graphicx}
\usepackage{hyperref}       % hyperlinks 
\hypersetup{
    colorlinks,
    linkcolor={red!90!black},
    citecolor={green!90!black},
    urlcolor={blue!30!black}
}
\usepackage{url}            % simple URL typesetting
\usepackage{booktabs}       % professional-quality tables
\usepackage{amsfonts}       % blackboard math symbols
\usepackage{xcolor}         % colors
\usepackage{amssymb}
\usepackage{makecell}
\newcommand{\etal}{\textit{et al}.}
\newcommand{\ie}{\textit{i}.\textit{e}.}
\newcommand{\eg}{\textit{e}.\textit{g}.}

\usepackage{pifont}

\usepackage{arydshln}
\usepackage{wrapfig}
\usepackage{multirow}

\usepackage[
backend=biber,
style=ieee,
citestyle=numeric-comp,
maxbibnames=3,
maxcitenames=3,
doi=false,isbn=false,url=false,eprint=false
]{biblatex}
\defbibheading{bibliography}{\begin{center}\textbf{6. REFERENCES}\end{center}}

\title{VISUAL SPEECH RECOGNITION FOR LANGUAGES WITH\\LIMITED LABELED DATA USING AUTOMATIC LABELS FROM WHISPER}
\name{Jeong Hun Yeo$^{1\dagger}$, Minsu Kim$^{1\dagger}$, Shinji Watanabe$^2$, Yong Man Ro$^{1*}$\thanks{$^\dagger$Equal Contribution. *Corresponding Author. This work was partially supported by the National Research Foundation of Korea (NRF) grant funded by the Korea government (MSIT) (No.~NRF-2022R1A2C2005529) and BK21 FOUR (Connected AI Education \& Research Program for Industry and Society Innovation, KAIST EE, No. 4120200113769).}} 
\address{$^1$Integrated Vision and Language Lab, KAIST, South Korea\\
$^2$Language Technologies Institute, Carnegie Mellon University, USA\\
\small{\texttt{\{sedne246,ms.k,ymro\}@kaist.ac.kr}, \,\,\texttt{swatanab@andrew.cmu.edu}}}

\begin{document}
\ninept
\maketitle
\begin{abstract}
This paper proposes a powerful Visual Speech Recognition (VSR) method for multiple languages, especially for low-resource languages that have a limited number of labeled data. Different from previous methods that tried to improve the VSR performance for the target language by using knowledge learned from other languages, we explore whether we can increase the amount of training data itself for the different languages without human intervention. To this end, we employ a Whisper model which can conduct both language identification and audio-based speech recognition. It serves to filter data of the desired languages and transcribe labels from the unannotated, multilingual audio-visual data pool. By comparing the performances of VSR models trained on automatic labels and the human-annotated labels, we show that we can achieve similar VSR performance to that of human-annotated labels even without utilizing human annotations. Through the automated labeling process, we label large-scale unlabeled multilingual databases, VoxCeleb2 and AVSpeech, producing 1,002 hours of data for four low VSR resource languages, French, Italian, Spanish, and Portuguese. With the automatic labels, we achieve new state-of-the-art performance on mTEDx in four languages, significantly surpassing the previous methods. The automatic labels are available online: \href{https://bit.ly/3Lajr6w}{bit.ly/3Lajr6w}
\end{abstract}

\vspace{-0.1cm}
\section{Introduction}
\vspace{-0.1cm}
These days, speech recognition technology \cite{prabhavalkar2023end,guo2021recent,li2022recent} stands out as one of the most widely adopted technologies integrated into the daily lives of individuals. Its importance is being emphasized along with the growth of conversational AI \cite{chatgpt}, as it is one of the basic communication channels between humans and AI. The main research stream for the technology is Audio-based Speech Recognition (ASR) \cite{hannun2014deep,amodei2016deep2,watanabe2017hybrid,gulati2020conformer,chang2023exploration} which utilizes audio modality to capture the spoken language. However, we can easily observe the technology fails to capture utterances accurately in noisy places, such as in a restaurant or conference hall. Different from the audio modality, since visual modality is not affected by auditory noises, a technology called Visual Speech Recognition (VSR) \cite{assael2016lipnet} is developed to mitigate the limitations of ASR. VSR, also referred to as lip reading, utilizes talking face video to predict the spoken language by focusing on modeling lip movements. As VSR systems only utilize visual information, they can be utilized for capturing speech in noisy environments or having a conversation with a person who cannot make a voice \cite{kim2023liptospeech,choi2023intelligible}. With its practical aspects, many research efforts \cite{afouras2018deep,petridis2018end,zhang2019spatio,zhao2020hearing,ma2021end,kim2022distinguishing,yeo2023multi,yeo2023akvsr,chang2023conformers,afouras2020asr,ivanko2022visual} have been made to advance the VSR technology. With these efforts, the current state-of-the-art VSR system \cite{ma2023auto} achieved a Word Error Rate (WER) of less than 15\% on a popular English VSR database, LRS2 \cite{chung2017lrs2}. By examining the performance of an early pioneering work \cite{chung2017lrs2}, which exhibited a WER of over 70\%, we can gauge the extent to which the technology has advanced.

%------------------------------------ Figure 1
%#################################################
\begin{figure}[t]
	\begin{minipage}[b]{1.0\linewidth}
		\centering
		\centerline{\includegraphics[width=8.5cm]{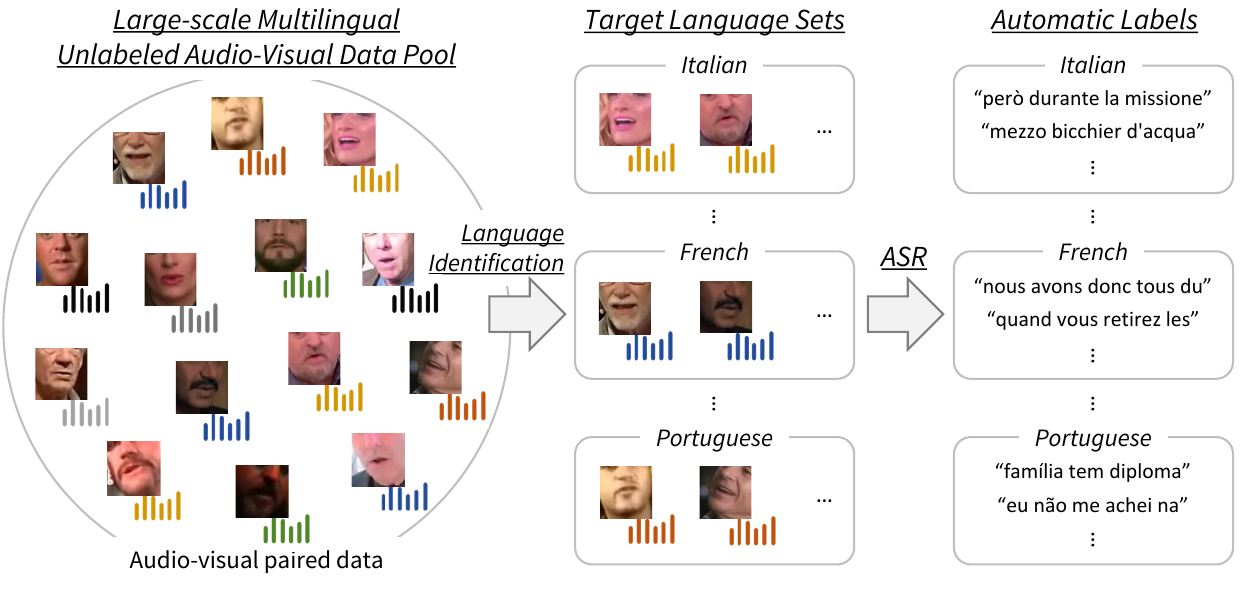}}
	\end{minipage}
	\vspace{-0.8cm}
	\caption{The conceptual illustration of automatic labeling process.}
	\label{fig:1}
	\vspace{-0.4cm}
\end{figure}
%##################################################

Despite the great development of VSR, the exploration of its performance for other languages than English has not well been made \cite{kim2023lmd}. Basically, this is because of the insufficient video-text data in different languages while the databases in English are steadily developed \cite{chung2017lrw,chung2017lrs2,afouras2018lrs3}. The publicly available English video-text data (\ie, LRW \cite{chung2017lrw}, LRS2 \cite{chung2017lrs2}, and LRS3 \cite{afouras2018lrs3}) construct over 870 hours while the labeled Italian data \cite{salesky2021mtedx} only compose about 50 hours \cite{ma2022multiple,kim2023lmd}. Since acquiring accurate text annotations for different languages is a burdensome task that requires expensive international human labor, previous approaches \cite{ma2022multiple,kim2023lmd,zinonos2023learning} tried to improve the VSR performances of low VSR resource languages by using the knowledge learned from high-resource language data or combined multilingual data. For example, \cite{kim2023lmd} proposed to learn general speech knowledge from rich English data and to learn language-specific knowledge from the target language data using a language-specific memory-augmented decoder. By combining the two learned knowledge, they achieved promising VSR performances even for low VSR resource languages.

Different from the previous approaches, we investigate whether it is possible to increase the amount of data itself for low VSR resource languages, without human intervention. Motivated from \cite{ma2023auto}, which utilized a pre-trained ASR model to transcribe unlabeled English video to increase the amount of labeled data, we investigate the performances of VSR models trained using automatically generated transcriptions for four low VSR resource languages, French (FR), Italian (IT), Spanish (ES), and Portuguese (PT). To this end, we bring two large-scale unlabeled multilingual audio-visual databases, VoxCeleb2 \cite{chung2018voxceleb2} and AVSpeech \cite{ephrat2018avspeech}. Since the databases compose different languages without annotations, we first perform language identification \cite{ambikairajah2011language} by using a pre-trained language classifier, to get language ID for each utterance. Then, for the utterances in the target language (\ie, FR, IT, ES, and PT), we apply a pre-trained ASR model to obtain the transcriptions for the unlabeled video. We first compare the effectiveness of these automatically labeled data with human-labeled data to analyze the validity of the automated labeling process, and show that we can achieve similar VSR performance with the human-labeled system even if we utilized the automatically labeled data only. Then, by merging the automatically labeled data (\ie, VoxCeleb2 and AVSpeech) with the human-labeled data (\ie, mTEDx), we train VSR models for each target language and evaluate them on a popular benchmark database, mTEDx \cite{salesky2021mtedx}.

The major contributions of this paper are as follows, 1) This is the first work exploring VSR for low VSR resource languages by using an automatic labeling process. By employing a Whisper model as both a pre-trained language identifier and an ASR model, we expand the amount of video-text data for low VSR resource languages (\ie, FR, IT, ES, and PT) by about 1,002 hours. 2) We set new state-of-the-art performances on benchmark databases in all four languages. 3) By analyzing the effectiveness of the automated labeling process compared to human labeling, we show that similar performance to human labeling can be achieved even if automated labeling is utilized. 4) We release the full automatic labels on \href{https://bit.ly/3Lajr6w}{bit.ly/3Lajr6w}.

\vspace{-0.1cm}
\section{Labeling for Low VSR Resource Languages}
\vspace{-0.1cm}
\subsection{Pre-trained Language Identifier and ASR Model}
To automatically produce the transcription for each target language from the unlabeled multilingual data pool, we require a pre-trained language identifier and an ASR model. Recently, Whisper \cite{radford2023whisper}, a powerful multilingual ASR model that can perform diverse tasks including language identification has been proposed. Since it showed significant multilingual ASR benchmark performances, and promising results in labeling the English dataset \cite{ma2023auto}, we utilize the Whisper model for our language identification and multilingual ASR. Specifically, we utilize the Whisper large model which is trained on about 680,000 hours of multilingual data with multitask supervision.

\subsection{Multilingual Audio-Visual Dataset}
\subsubsection{Labeled Multilingual Dataset}
Multilingual TEDx (mTEDx) \cite{salesky2021mtedx} is a multilingual corpus for speech recognition and translation. It is a human-annotated dataset as the transcriptions for the dataset are created by volunteers. Since the dataset only includes speech audio and transcriptions, the videos are acquired by downloading them online through the provided links. We follow \cite{ma2022multiple,kim2023lmd} to remove the video clips not contain a speaker or are unavailable. This data is basically utilized to evaluate the VSR performances in four different languages, French (FR), Italian (IT), Spanish (ES), and Portuguese (PT). The data statistics for each language are presented in Table \ref{table:1}. The abbreviation $\text{MT}$ represents mTEDx and its subscript represents the language ID.

\subsubsection{Unlabeled Multilingual Dataset}
We utilize two large-scale unlabeled multilingual audio-visual databases, VoxCeleb2 \cite{chung2018voxceleb2} and AVSpeech \cite{ephrat2018avspeech}. VoxCeleb2 \cite{chung2018voxceleb2} contains about 1.1M multilingual utterances from 6,112 speakers with 2,442 hours of unlabeled videos. Among them, about 1,116 hours of videos are non-English \cite{shi2021learning}. AVSpeech \cite{ephrat2018avspeech} contains about 4,700 hours of unlabeled multilingual videos of approximately 150,000 speakers. About 3,400 hours of videos are in other languages than English. Therefore, by merging the two databases, we can build roughly 4,500 hours of unlabeled multilingual audio-visual data pool after removing English data. From the constructed data pool, we detect and label the target language transcriptions. We use the abbreviations VC for VoxCeleb2 and AV for AVSpeech when referring to the respective datasets.

%########### Table 1 #############
\input{Table/Table1.tex}
%#################################

\vspace{-0.2cm}
\subsection{Automated Labeling and Analyzing the Effectiveness}
With the pre-trained language identifier and ASR model, obtaining labels from the constructed unlabeled data pool is straightforward. First, we perform language identification using Whisper, and the data predicted as the target languages, FR, IT, ES, and PT, are kept only. Then, Whisper is applied again with the predicted language labels to infer the transcription of the input speech. Throughout this automated labeling process, we exclude data where the predicted characters do not belong to the target language.

Our primary consideration stems from the fact that no technology is flawless. It is imperative to acknowledge the existence of data characterized by inaccurate language and sentences that have been predicted incorrectly. In order to assess the impact of model prediction errors on diminishing the quantity of reliable data and influencing the final developed VSR model, we initiate by subjecting the mTEDx \cite{salesky2021mtedx} dataset to the automated labeling process, simulating a scenario where it lacks any labels. Specifically, we discard the transcriptions of mTEDx and mix training data from all languages to simulate that there is no information for language IDs and transcription. Then onto this simulated multilingual unlabeled audio-visual data pool, we apply the automated labeling process (\ie, language identification and ASR). By doing this, we can figure out how much data will be diminished by the incorrect labeling process and how this data influences VSR performance, especially compared to human-labeled data. 

%########### Table 2 #############
\input{Table/Table2.tex}
%#################################

%########### Table 3 #############
\input{Table/Table3.tex}
%#################################

Table \ref{table:2} shows the data amount comparisons between the original human-labeled data and the automatically labeled data. We observed that the automated process preserves a significant portion of the data and generates comparable amounts with the human-labeled data across languages, with the exception of Spanish. For FR, IT, and PT, we can preserve more than 99.0\% of training data compared to the original human-labeled data. While inspecting the Spanish dataset, we observed instances where some Spanish data were classified as Galician (GL), a Spanish variant. For the purposes of this paper, we exclusively employ data classified as Spanish (ES) only. Moreover, we also measure the WER to check how much the prediction of the ASR model (\ie, Whisper) deviates from the human-labeled transcription. The measured WERs for the four languages are from about 7\% to 12\%, showing that the ASR model makes errors when labeling the transcriptions. 

To evaluate how the automatic labels affect the final VSR performance, we train a total of eight VSR models for four languages using both the original human-labeled data and the aforementioned automatically labeled data. Then, we evaluate the trained models on the test splits of the mTEDx dataset. Table \ref{table:3} shows the evaluation results. We report both WER and Character Error Rate (CER). When we compare the performances between human-labeled and automatically labeled models, we can find that the models trained using automatically labeled data achieve similar performances to the models trained using the original data. Furthermore, it's worth noting that in the case of Spanish, we can achieve better performance than the original human-labeled data. Finally, it becomes evident that a labeling error of roughly lower than 12\% WER is acceptable for training a VSR model with automatic labels.

\vspace{-0.1cm}
\section{Experiments}
\vspace{-0.2cm}

\subsection{VSR Model}
For the VSR model, we employ an encoder-decoder model \cite{ma2021end} with joint CTC/attention training \cite{watanabe2017hybrid}. For its front-end, ResNet-18 \cite{he2016deep} with a 3D CNN stem layer is utilized \cite{petridis2018end}. The model consists of a 12-layer Conformer \cite{gulati2020conformer} encoder and a 6-layer transformer \cite{vaswani2017attention} decoder. Each layer is comprised of 12 attention heads, feed-forward dimensions of 3,072, and latent dimensions of 768. We initialize the weights of the model with a pre-trained English lip reading model \cite{ma2023auto}, following \cite{kim2023lmd}. For the decoding, we use beam search with a beam width of 40, and no external language model is utilized.

\vspace{-0.2cm}
\subsection{Implementation details}
We employ the pre-trained multilingual ASR model, Whisper \cite{radford2023whisper}, open-sourced in the Hugging Face community with the model card name "openai/whisper-large-v2" for both identifying four languages (\ie, FR, IT, ES, and PT) and generating transcriptions.

%########### Table 4 #############
\input{Table/Table4.tex}
%#################################

For the video pre-processing, we first detect and extract the facial landmarks using \cite{deng2020retinaface}, and then we align the face to the reference face frame following \cite{ma2023auto}. Finally, the mouth region is cropped by using a bounding box with a size of $96 \times 96$. The data augmentation of horizontal flipping, random cropping, and time masking are utilized. For the transcription, we use SentencePiece \cite{kudo2018sentencepiece} subword units with a vocabulary size of 1,000 in all four languages. 

We train the VSR model for 40 epochs with the AdamW \cite{loshchilov2018decoupled,kingma2014adam} optimizer. A cosine learning rate scheduler with a warm-up for 5 epochs is utilized. The peak learning rate is set to $4 \times 10^{-4}$. We set the max frame length of video clips as 600. For training on mTEDx \cite{salesky2021mtedx} and VoxCeleb2 \cite{chung2018voxceleb2}, the number of frames in each batch is limited to 1800, and 2 RTX3090 GPUs are used. When we additionally use AVSpeech \cite{ephrat2018avspeech} for training the model, the number of frames in each batch is limited to 1200, and 4 RTX3090 GPUs are used.

\vspace{-0.1cm}
\subsection{Automatic Labeling Results}
By applying the automated labeling process to the unlabeled multilingual audio-visual data pool, we get a total of 1,002 hours of additional training data for the four languages. By merging the human-labeled data composed of 285 hours with the automatically labeled data, we build a total of 1,287 hours of training data. The details of the training data for each language and dataset are shown in Table \ref{table:1}.

\vspace{-0.1cm}
\subsection{Experimental Results}
With the automatically labeled data, we train VSR models for each of four languages, FR, IT, ES, and PT. We compare the VSR performances with the previous state-of-the-art VSR methods.

\vspace{-0.2cm}
\subsubsection{Comparison with the state-of-the-art methods}
Table \ref{table:4} shows the performance comparison results on mTEDx databases. As some methods \cite{ma2022multiple,zinonos2023learning} only report WER or CER, to compare with them, we report both WER and CER. By employing the automatically labeled data in the training of the VSR models, the proposed methods outperform the previous methods \cite{ma2022multiple,zinonos2023learning,kim2023lmd} in all four languages. Notably, for the languages ES and PT, there are much larger amounts of labeled data created compared to the size of the human-labeled data (about 4 times larger). This offers further enhancement of the VSR performance compared to the languages FR and IT. For example, in Portuguese VSR, the model achieves 47.89\% WER which is better than the previous state-of-the-art \cite{kim2023lmd} with about 10.9\% WER improvement. From the experimental results, we clearly show the effectiveness of the proposed method by achieving new state-of-the-art performances on benchmark databases in all four languages. The proposed method improves over 10\% WER in two languages (\ie, ES and PT), while previous methods competed within 3\% WERs and CERs ranges. Please note that as the proposed automatic labeling can provide rich training data, it can be jointly employed with the advanced training methods of the previous methods \cite{ma2022multiple,zinonos2023learning,kim2023lmd}, to further improve the VSR performance.

\vspace{-0.2cm}
\subsubsection{Impact of the size of automatically labeled data} 
In order to check the impact of the size of automatically labeled data from VoxCeleb2 and AVSpeech, we perform an ablation study by training the VSR models with different amounts of automatically labeled data. To this end, we train the models with human-labeled data only (MT), with automatic labels from VoxCeleb2 (MT+VC), and finally with the full automatic labels including that of AVSpeech (MT+VC+AV). Table \ref{table:5} shows the amounts of training data and their performances for each language. The model trained on human-labeled data only achieves 65.25\%, 60.40\%, 59.91\%, and 59.45\% WERs in FR, IT, ES, and PT, respectively. When we add automatic labels of VoxCeleb2 for training in each language, the performances for all languages are improved. By examining the results, we can find the correlation between the total training hours and the performances. In French (FR), Italian (IT), and Spanish (ES), the training hours are increased by more than approximately 40 hours, and the performances are also improved by about 4\% WERs or more. In contrast, the added training data for Portuguese (PT) is only 9 hours by using VoxCeleb2, and the reduction of WER is 0.63\%, which is a bit marginal. When we utilize both VoxCeleb2 and AVSpeech, we accomplish the best WERs for all languages. Especially, the performance for PT is greatly improved since we can bring roughly 330 hours of automatically labeled data from AVSpeech.

\vspace{-0.1cm}
\section{Relation to Prior Work}
\vspace{-0.1cm}
There are numerous efforts in VSR research, especially in English \cite{afouras2018deep,petridis2018end,zhang2019spatio,zhao2020hearing,ma2021end,kim2022distinguishing,yeo2023multi,yeo2023akvsr,chang2023conformers,afouras2020asr,ivanko2022visual,shi2021learning}. By employing Conformer in VSR architecture, \cite{ma2021end} showed promising VSR performances. Recently, \cite{ma2023auto} proposed to increase the visual-text data by automatically labeling unlabeled English audio-visual data, and achieved significant VSR performance. However, they were all focused on high-resource language VSR (\eg, English and Mandarin), leaving uncertainty regarding the applicability to other languages, especially those with limited VSR resources. Only a few recent approaches \cite{ma2022multiple,kim2023lmd,zinonos2023learning} explored the low-resource language problem in VSR. Since constructing multilingual visual-text data is a burdensome task that requires international human intervention, previous works focused on transferring the general knowledge extracted from the high-resource language data to mitigate the low-resource problem. Different from these approaches, we try to directly address the low VSR resource problem by increasing the labeled data through automatically labeling the multilingual data. The experimental results demonstrate that increasing the data size itself for low VSR resource languages is another important factor in advancing multilingual VSR systems. Finally, since the proposed method can be jointly integrated with existing learning-based approaches to build powerful multilingual VSR systems, we emphasize the significance of our exploration in the automated labeling process for multilingual data.

%########### Table 5 #############
\input{Table/Table5.tex}
%#################################

\vspace{-0.1cm}
\section{Conclusion}
\vspace{-0.1cm}
In this paper, we proposed a powerful Visual Speech Recognition (VSR) framework especially for low VSR resource languages, French, Italian, Spanish, and Portuguese. We aimed to handle the issue of insufficient label problems without a human-labeling process. To achieve this, we utilized a pre-trained Whisper model to detect language IDs and generate transcriptions from a large-scale unlabeled multilingual audio-visual data pool. We analyzed the effectiveness of the automatic labeling by comparing it with the human-labeled data, and we showed that the human-labeling process could be replaced by the automated labeling process. With the extended training data, we achieved new state-of-the-art performances with large improvements from the previous methods in all languages, showing the importance of addressing the low VSR resource language problem.

% References should be produced using the bibtex program from suitable
% BiBTeX files (here: strings, refs, manuals). The IEEEbib.bst bibliography
% style file from IEEE produces unsorted bibliography list.
% -------------------------------------------------------------------------

% {\small
% \bibliographystyle{IEEEbib}
% \bibliography{refs}
% }

\printbibliography[heading=bibliography]
\end{document}

%% file: Table/Table1.tex
\begin{table}[t]
	\renewcommand{\arraystretch}{1.2}
	\renewcommand{\tabcolsep}{3mm}
\vspace{-0.2cm}
\caption{Details of the VSR training data including automatic labels. Abbreviations - MT: mTEDx, VC: VoxCeleb2, AV: AVSpeech.}
\vspace{0.1cm}
\centering
\resizebox{0.9\linewidth}{!}{
\begin{tabular}{ccccc}
\Xhline{3\arrayrulewidth}
\textbf{Dataset} & \textbf{\makecell{Human-labeled \\ \# of Video}} & \textbf{\makecell{Auto-labeled \\ \# of Video}} & \textbf{Hours} &\textbf{\makecell{Total\\Hours}} \\ \hline
$\text{MT}_{\text{FR}}$ & 58,426  & - & 85 & \multirow{3}{*}{331}  \\
$\text{VC}_{\text{FR}}$ & -  & 66,943 & 124 & \\
$\text{AV}_{\text{FR}}$ & -  & 69,020 & 122 & \\ 
\hline
$\text{MT}_{\text{IT}}$ & 26,108  & - & 46 & \multirow{3}{*}{152}\\
$\text{VC}_{\text{IT}}$ & - & 19,261 & 38 & \\ 
$\text{AV}_{\text{IT}}$ & - & 38,227 & 68 & \\
\hline

$\text{MT}_{\text{ES}}$ & 44,532  & - & 72 & \multirow{3}{*}{384}\\
$\text{VC}_{\text{ES}}$ & - & 22,682 & 42 & \\
$\text{AV}_{\text{ES}}$ & - & 151,173 & 270 & \\

\hline
$\text{MT}_{\text{PT}}$ & 52,058 & - & 82 & \multirow{3}{*}{420}\\
$\text{VC}_{\text{PT}}$ & - & 4,843 & 9 & \\
$\text{AV}_{\text{PT}}$ & - & 176,601 & 329 & \\ 
\Xhline{3\arrayrulewidth}
\end{tabular}}
\label{table:1}
\vspace{-0.4cm}
\end{table}

%% file: Table/Table2.tex
\begin{table}[t]
	\renewcommand{\arraystretch}{1.2}
	\renewcommand{\tabcolsep}{2mm}
\vspace{-0.2cm}
\caption{Comparison between human-labeled data and automated labeled data on mTEDx (MT). Subscript refers to language ID.}
\vspace{0.1cm}
\centering
\resizebox{0.99\linewidth}{!}{
\begin{tabular}{ccccc}
\Xhline{3\arrayrulewidth}
\textbf{Data} & \textbf{\makecell{Human-labeled \\ \# of Video}} & \textbf{\makecell{Auto-labeled \\ \# of Video}} & \textbf{\makecell{Preserved\\ Ratio (\%)}} & \textbf{
\makecell{Labeling \\ WER (\%)}} \\ \hline
$\text{MT}_\text{FR}$ & 58,426 & 58,222 & 99.7 & 11.61 \\ 
$\text{MT}_\text{IT}$ & 26,108 & 25,898 & 99.2 & 7.55 \\ 
$\text{MT}_\text{ES}$ & 44,532 & 37,853 & 85.0 & 9.18 \\
$\text{MT}_\text{PT}$ & 52,058 & 51,555 & 99.0 & 11.69 \\ 
\Xhline{3\arrayrulewidth}
\end{tabular}}
\label{table:2}
\vspace{-0.4cm}
\end{table}

%% file: Table/Table3.tex
\begin{table}[t]
	\renewcommand{\arraystretch}{1.2}
	\renewcommand{\tabcolsep}{3.8mm}
\caption{VSR performances trained using human-labeled data and auto-labeled data on mTEDx (MT). Subscript refers to language ID.}
\vspace{0.1cm}
\centering
\resizebox{0.99\linewidth}{!}{
\begin{tabular}{ccccc}
\Xhline{3\arrayrulewidth}
\multirow{2}{*}{\textbf{Data}} & \multicolumn{2}{c}{\textbf{Human-labeled}} & \multicolumn{2}{c}{\textbf{Auto-labeled}} \\ \cmidrule(l{2pt}r{2pt}){2-3}\cmidrule(l{2pt}r{2pt}){4-5}
& \textbf{WER(\%)} & \textbf{CER(\%)} & \textbf{WER(\%)} & \textbf{CER(\%)} \\ \hline
$\text{MT}_\text{FR}$ & 65.25 & 42.93 & 69.33 & 49.22 \\ 
$\text{MT}_\text{IT}$ & 60.40 & 33.53 & 59.17 & 34.83 \\ 
$\text{MT}_\text{ES}$ & 59.91 & 33.93 & 58.57 & 33.64 \\
$\text{MT}_\text{PT}$ & 59.45 & 37.49 & 59.70 & 36.91 \\ 
\Xhline{3\arrayrulewidth}
\end{tabular}} 
\label{table:3}
\vspace{-0.4cm}
\end{table}

%% file: Table/Table4.tex
\begin{table*}[h]
  \renewcommand{\arraystretch}{1.3}
  \renewcommand{\tabcolsep}{2.7mm}
  \centering
  \caption{VSR results comparisons on mTEDx dataset (FR, IT, ES, and PT). Some works only report WER or CER, so we report both WER and CER to compare with them. CM represents CMU-MOSEAS \cite{zadeh2020cmu} which is not publicly available. Pre-training set size refers to the size of English video-text or Multilingual audio-visual data for pre-training the model before fine-tuning it on the target language data.}
  \vspace{0.1cm}
  \resizebox{0.999\linewidth}{!}{
  \begin{tabular}{cccccccc}
    \Xhline{3\arrayrulewidth}
    \textbf{Method} & \textbf{Language} & \textbf{Target Language Dataset} & \makecell{\textbf{Pre-training} \\ \textbf{Set Size (hours)}} & \makecell{\textbf{Target Language} \\ \textbf{Set Size (hours)}} & \textbf{External LM} & \textbf{WER(\%)} & \textbf{CER(\%)} \\ \hline

    Ma \etal~\cite{ma2022multiple} & FR & $\text{MT}_{\text{FR}} + \text{CM}_{\text{FR}}$ & 1,459 & 100 & \checkmark & 66.20 & - \\
    Zinonos \etal~\cite{zinonos2023learning} & FR  & $\text{MT}_{\text{FR}}$ & 1,333 & 72 & - & - & 43.30  \\
    Kim \etal~\cite{kim2023lmd} & FR & $\text{MT}_{\text{FR}}$ & 3,448 & 85 & - & 64.92 & 42.69 \\ \hdashline
    \textbf{Proposed Method} & FR & $\text{MT}_{\text{FR}} + \text{VC}_{\text{FR}} + \text{AV}_{\text{FR}}$ & 3,448 & 331 & - & \textbf{58.30} & \textbf{37.85} \\
    \hline

    Ma \etal~\cite{ma2022multiple} & IT & $\text{MT}_{\text{IT}}$  & 1,459 & 46 & \checkmark & 57.40 & - \\
    Zinonos \etal~\cite{zinonos2023learning} & IT & $\text{MT}_{\text{IT}}$ & 1,333  & 40 & - & - & 35.10  \\
    Kim \etal~\cite{kim2023lmd} & IT & $\text{MT}_{\text{IT}}$ & 3,448 & 46 & - & 59.74 & 33.30 \\ \hdashline
    \textbf{Proposed Method} & IT & $\text{MT}_{\text{IT}} + \text{VC}_{\text{IT}} + \text{AV}_{\text{IT}}$ & 3,448 & 152 & - & \textbf{51.79} & \textbf{29.83}\\
    \hline

    Ma \etal~\cite{ma2022multiple} & ES & $\text{MT}_{\text{ES}} + \text{CM}_{\text{ES}}$ & 1,459 & 87 & \checkmark & 56.30 & - \\
    Zinonos \etal~\cite{zinonos2023learning} & ES & $\text{MT}_{\text{ES}}$ & 1,333 & 62 & - & - & 32.80  \\
    Kim \etal~\cite{kim2023lmd} & ES & $\text{MT}_{\text{ES}}$ & 3,448 & 72 & - & 56.96 & 32.28 \\ \hdashline
    \textbf{Proposed Method} & ES & $\text{MT}_{\text{ES}} + \text{VC}_{\text{ES}} + \text{AV}_{\text{ES}}$ & 3,448 & 384 & - & \textbf{45.71} & \textbf{25.90} \\
    \hline

    Ma \etal~\cite{ma2022multiple} & PT & $\text{MT}_{\text{PT}} + \text{CM}_{\text{PT}}$ & 1,459 & 99 & \checkmark & 61.50 & - \\
    Zinonos \etal~\cite{zinonos2023learning} & PT & $\text{MT}_{\text{PT}}$ & 1,333 & 73 & - & - & 38.70  \\
    Kim \etal~\cite{kim2023lmd} & PT & $\text{MT}_{\text{PT}}$ & 3,448 & 82 & - & 58.57 & 37.68 \\ \hdashline
    \textbf{Proposed Method} & PT & $\text{MT}_{\text{PT}} + \text{VC}_{\text{PT}} + \text{AV}_{\text{PT}}$ & 3,448 & 420 & -  & \textbf{47.89} & \textbf{29.74} \\    

    \Xhline{3\arrayrulewidth}
  \end{tabular}}
  \label{table:4}
\vspace{-0.5cm}
\end{table*}

%% file: Table/Table5.tex
\begin{table}[t]
	\renewcommand{\arraystretch}{1.2}
	\renewcommand{\tabcolsep}{2.3mm}
\vspace{-0.2cm}
\caption{Ablation study on automatically labeled data amounts. Abbreviations - MT: mTEDx, VC: VoxCeleb2, AV: AVSpeech.}
\vspace{0.1cm}
\centering
\resizebox{0.99\linewidth}{!}{
\begin{tabular}{ccccc}
\Xhline{3\arrayrulewidth}
\textbf{Datasets} & \textbf{Language} & \textbf{\makecell{Total \\ Hours}} & \textbf{WER(\%)} & \textbf{CER(\%)} \\ \hline
 $\text{MT}_{\text{FR}}$  & FR & 85 & 65.25 & 42.93 \\ 
$\text{MT}_{\text{FR}} + \text{VC}_{\text{FR}}$ & FR & 209 & 60.61 & 38.79 \\ 
 $\text{MT}_{\text{FR}} + \text{VC}_{\text{FR}} + \text{AV}_{\text{FR}}$  & FR & 331 & \textbf{58.30} & \textbf{37.85} \\ 
 \hline
 $\text{MT}_{\text{IT}}$  & IT & 46 & 60.40 & 33.53 \\ 
$\text{MT}_{\text{IT}} + \text{VC}_{\text{IT}}$ & IT & 84 & 56.48 & 31.52 \\ 
 $\text{MT}_{\text{IT}} + \text{VC}_{\text{IT}} + \text{AV}_{\text{IT}}$  & IT & 152 & \textbf{51.79} & \textbf{29.83} \\ 
 \hline
  $\text{MT}_{\text{ES}}$  & ES & 72 & 59.91 & 33.93 \\ 
$\text{MT}_{\text{ES}} + \text{VC}_{\text{ES}}$ & ES & 114 & 54.05 & 31.03 \\ 
 $\text{MT}_{\text{ES}} + \text{VC}_{\text{ES}} + \text{AV}_{\text{ES}}$  & ES & 384 & \textbf{45.71} & \textbf{25.90} \\ 
 \hline

  $\text{MT}_{\text{PT}}$  & PT & 82 & 59.45 & 37.49 \\ 
$\text{MT}_{\text{PT}} + \text{VC}_{\text{PT}}$ & PT & 91 & 58.82 & 36.96 \\ 
 $\text{MT}_{\text{PT}} + \text{VC}_{\text{PT}} + \text{AV}_{\text{PT}}$  & PT & 420 & \textbf{47.89} & \textbf{29.74} \\ 

\Xhline{3\arrayrulewidth}
\end{tabular}}
\label{table:5}
\vspace{-0.5cm}
\end{table}